\documentclass[lettersize,twoside,journal]{IEEEtran}
\usepackage{amsmath,amsfonts}
\usepackage{array}
\usepackage{textcomp}
\usepackage{stfloats}
\usepackage{verbatim}
\usepackage{graphicx} 
\usepackage{subfigure}

\def\BibTeX{{\rm B\kern-.05em{\sc i\kern-.025em b}\kern-.08em
    T\kern-.1667em\lower.7ex\hbox{E}\kern-.125emX}}

\usepackage{algorithm}
\usepackage{algorithmicx}
\usepackage{algpseudocode}
\usepackage{amssymb}
\usepackage[thmmarks, amsmath, thref]{ntheorem}
\usepackage{extarrows}
\usepackage{epstopdf}
\usepackage{setspace}
\usepackage{cite}
\usepackage{authblk}
\usepackage{booktabs}
\usepackage{multirow}
\usepackage{makecell}
\usepackage{bm}
\usepackage{orcidlink}

\hypersetup{hidelinks}
\hypersetup{
    colorlinks=true,       %
    linkcolor=blue,         %
    urlcolor=red,           
    citecolor=green,        %
    filecolor=magenta       %
}
\begin{document}
\title{HiLo: Learning Whole-Body Human-like Locomotion with Motion Tracking Controller}

\author{Qiyuan Zhang$^\ast$, Chenfan Weng$^\ast$, Guanwu Li,
Fulai He, and Yusheng Cai$^\dagger$
\thanks{$^\ast$Equal contribution. $^\dagger$Corresponding author. All authors are with the Department of Research and Development, Fourier Intelligence, Pudong, Shanghai 200120, China. {\tt\small $\{$qiyuan.zhang, chenfan.weng, guanwu.li, fulai.he, yusheng.cai$\}$@fftai.com}

This work has been submitted to the IEEE for possible publication. Copyright may be transferred without notice, after which this version may no longer be accessible.}
}

\maketitle

\begin{abstract}
Deep Reinforcement Learning (RL) has emerged as a promising method to develop humanoid robot locomotion controllers. Despite the robust and stable locomotion demonstrated by previous RL controllers, their behavior often lacks the natural and agile motion patterns necessary for human-centric scenarios. In this work, we propose HiLo (human-like locomotion with motion tracking), an effective framework designed to learn RL policies that perform human-like locomotion. The primary challenges of human-like locomotion are complex reward engineering and domain randomization. HiLo overcomes these issues by developing an RL-based motion tracking controller and simple domain randomization through random force injection and action delay. Within the framework of HiLo, the whole-body control problem can be decomposed into two components: One part is solved using an open-loop control method, while the residual part is addressed with RL policies. A distributional value function is also implemented to stabilize the training process by improving the estimation of cumulative rewards under perturbed dynamics. Our experiments demonstrate that the motion tracking controller trained using HiLo can perform natural and agile human-like locomotion while exhibiting resilience to external disturbances in real-world systems. Furthermore, we show that the motion patterns of humanoid robots can be adapted through the residual mechanism without fine-tuning, allowing quick adjustments to task requirements.

Index Terms - Whole-Body Control, Human-like Locomotion, Reinforcement Learning, Humanoid Robots, Sim-to-Real

\end{abstract}

\section{INTRODUCTION}
\IEEEPARstart{H}{umanoid} robotics has intrigued many research institutes and companies due to its potential in general-purpose AI \cite{liu2024aligning, vernon2024importance} and diverse applications \cite{li2024reinforcement, radosavovic2024real}. These human-sized robots have advantages in tasks that require dexterity and adaptability in human-centric settings. Whole-body locomotion control is a fundamental component that allows these robots to perform effectively. Traditional robot control approaches, such as model predictive control (MPC) \cite{garcia1989model, farshidian2017real}, have demonstrated impressive mobility through precise mathematical models and predefined motion planning. However, these methods often struggle with robustness and generalization in unknown or dynamically changing environments. Moreover, they require significant expertise for dynamics modeling and maintenance, which limits their broader applicability. Recently, Reinforcement Learning (RL) has emerged as a promising solution for legged robots \cite{hwangbo2019learning}. RL allows robots to acquire complex motor skills through interactions with their environment, showing potential to overcome the limitations of traditional methods.

Despite recent advances in RL for humanoid locomotion \cite{tong2024advancements}, the high dimensionality of humanoid robots poses significant challenges for RL policies to acquire human-like motion patterns required for human-centric scenarios. This is particularly evident in environments that require human interaction and cooperation \cite{semeraro2023human}, such as healthcare settings and collaborative tasks. Prior RL methods commonly struggle with reward engineering and domain randomization, which are critical components of learning natural and agile locomotion. Designing a reward function that accurately captures the desired behavior can be a laborious and time-consuming process. Given the complexity and inherent instability of humanoid dynamics, a well-designed reward function must also balance multiple objectives, including stability, energy efficiency, and natural movement. Domain randomization \cite{andrychowicz2020learning}, on the other hand, requires the creation of training environments with diverse dynamics properties to ensure that the learned RL policy adapts to real-world systems. This requires a deep understanding of robot dynamics, potential discrepancy in real-world conditions, and the tasks at hand. These challenges have significantly limited the practicality and effectiveness of current RL methods to achieve truly human-like locomotion.

In this work, we present HiLo (Human-like Locomotion with Motion Tracking), an effective RL policy learning framework for human-like locomotion. HiLo trains a motion tracking controller that optimizes a mimic reward, penalizing deviations from a reference walking motion. This approach simplifies the reward design and provides a clear signal for control. The motion tracking controller consists of two components: an open-loop control part based on the target reference and a residual component that is solved by an RL policy. This residual mechanism allows the RL policy to make corrections for deviations instead of having to learn locomotion from scratch. Additionally, the decomposition of the motion tracking controller offers flexibility for modifying motion patterns without fine-tuning. To enhance training stability and efficiency, HiLo utilizes a distributional value function to capture perturbed dynamics information, leading to a more accurate estimation of cumulative rewards. Furthermore, HiLo also integrates extended random force injection \cite{campanaro2024learning} with action delay to replace complex dynamics perturbation techniques. This integration improves policies' robustness and facilitates transfer from simulations to real-world applications. In summary, our contributions are four-fold:

\begin{itemize}
  \item \textbf{Simple yet effective domain randomization for humanoid robots.} Unlike traditional methods that randomize various dynamics properties, we randomize dynamics by injecting random forces into motor torques and introducing action delays to simulate the nonlinearity of the actuation motors. This approach relies on a single hyperparameter and requires fewer than 20 lines of code.
  
  \item \textbf{Distributional value estimation to accelerate the training of the motion tracking controller.} Environmental uncertainty from domain randomization limits traditional RL methods that rely on expected value estimation. By adopting a distributional value function, we capture the full range of possible returns, resulting in a 2.5 times faster convergence.
  
  \item \textbf{Rapid adaptation of humanoid motion patterns without fine-tuning.} The motion tracking controller utilizes a combination of open-loop control and an RL policy. This approach allows for a variety of movements by adjusting the reference motion. Although the learned controller is optimized for the forward walking locomotion task, it can perform diverse motion patterns with zero-shot transfer.
  
  \item \textbf{Comprehensive experiments conducted in high-fidelity simulation and real-world systems.} We show the feasibility of leveraging the RL-based motion tracking controller to produce human-like locomotion for humanoid robots. 
\end{itemize}
The rest of this paper is organized as follows: Section \ref{related work} reviews related work in humanoid locomotion and domain randomization. Section \ref{prelimin} outlined the foundational concepts necessary for the proposed method. Section \ref{metho} details the HiLo framework. Section \ref{reseult} presents our experimental setup and results, demonstrating the effectiveness of our method in both simulated and real-world scenarios. Finally, Section \ref{concul} concludes the work and discusses future directions.

\section{RELATED WORK}\label{related work}
\textbf{Motion Tracking.} Traditional RL methods for humanoid robots \cite{radosavovic2024real, siekmann2021sim} are highly dependent on hand-crafted reward signals to promote natural gaits, which require significant human insight and domain expertise. However, recent work has proposed the incorporation of RL with reference motion data to reduce human effort. For example, DeepMimic \cite{peng2018deepmimic} and the perpetual humanoid controller (PHC) \cite{luo2023perpetual} have shown impressive results in generating physically plausible character animations by precisely tracking reference motion data. In addition, Peng et al. \cite{peng2021amp, peng2022ase} have incorporated RL with adversarial imitation learning to produce lifelike behaviors. Despite these advancements in physics-based character animation, there is limited evidence supporting the feasibility and effectiveness of motion tracking in humanoid robots. Notable exceptions include \cite{cheng2024expressive}, where RL policies are trained to track the reference motion of the upper body. He et al. \cite{he2024hover} proposed integrating various control modes into a single policy to track user-specified reference motion. Another study \cite{zhang2024whole} integrates AMP rewards with traditional gait rewards to promote natural full-body motions. In our work, we remove the need for separate gait generation and utilize reference motion as both phase information and reward signals to train a motion tracking controller that achieves human-like whole-body control.

\textbf{Domain Randomization.} Recent advances in RL-based humanoid locomotion \cite{radosavovic2024real,radosavovic2024learning} have been substantially propelled by domain randomization, which is crucial to enhancing the robustness and generalization of RL policies. Traditional domain randomization \cite{andrychowicz2020learning,radosavovic2024real} involves varying terrain types, robot dynamics, and actuator dynamics. This variability helps RL agents learn policies that are more adaptable to real-world conditions, thereby reducing the sim-to-real gap. However, these approaches need an exhaustive identification of the system and the identification of the dynamics distribution. As an alternative to complex domain randomization methods, \cite{valassakis2020crossing} demonstrated impressive sim-to-real performance in manipulation tasks using a simple randomized force injection (RFI) strategy, which emulates dynamics randomization by perturbing the dynamics of the system with randomized forces. ERFI \cite{campanaro2024learning} extended RFI by introducing an episodic actuation offset to perturb the dynamics of quadrupedal robots during training. Although simple RFI and ERFI have proven effective for manipulators and quadrupedal robots, their effectiveness has not been clearly demonstrated in humanoid robots. The substantial mass of a humanoid's upper body forces the actuation motors located in the lower body to operate under a loaded condition. This results in a noticeable delay in response to control commands, which hinders the effectiveness of ERFI in humanoid applications. To address this issue, we integrate ERFI with action delay to effectively perturb the dynamics of humanoid robots. Our experiments show that the RL-based controller trained within this simple domain randomization can be applied to real humanoid robots with zero-shot transfer.

\section{Preliminaries}\label{prelimin}
\subsection{Reinforcement Learning Framework}
To develop an RL-based motion tracking controller for human-like locomotion tasks, we formulate this problem within the framework of goal-conditioned reinforcement learning \cite{sutton2018reinforcement}. In this framework, an RL agent interacts with the environment according to a policy $\pi$. At each time step $t$, the agent receives a state $s_t\in \mathcal{S}$ and a target goal $g_t\in\mathcal{G}$ from the environment. The agent then samples an action $a_t\in\mathcal{A}$ from the policy $\pi$, with $\pi(a_t|s_t, g_t)$ denoting the probability of selecting action $a_t$ in $s_t$ and $g_t$. According to the environmental dynamics $p(s_{t+1}|s_t,a_t)$, the environment transits from $s_t$ to $s_{t+1}$, and the agent receives a reward $r_t:=r(s_t,a_t,g_t)$. The goal of the RL agent is to find an optimal policy $\pi^\ast$ to maximize the expected return:
\begin{equation}
    J=\mathbb{E}_{a_t\sim \pi(\cdot|s_t), s_{t+1}\sim p(\cdot|s_t,a_t)}\left[\sum^T_{t=0}\gamma^t r_t\right]
\end{equation}
where $\gamma\in (0,1]$ is the discount factor and $T$ is the episode length.

Let $R^\pi_t:=\sum^\infty_{t^\prime=t}\gamma^{t^\prime-t}r_{t^\prime}$ denote the discounted return from $s_t$ given a policy $\pi$ at time $t$. The expectation of $\mathcal{R}^\pi_t$ is expressed as 
\begin{equation}
    V^\pi(s):=\mathbb{E}_{a\sim\pi(\cdot|s)}\left[Q^\pi(s,a)\right]=\mathbb{E}\left[\mathcal{R}_t^\pi|s_t=s\right],
    \nonumber
\end{equation}
where $Q^\pi(s,a):=\mathbb{E}\left[\mathcal{R}_t^\pi|s_t=s,a_t=a\right]$. The advantage function is denoted as $A^\pi(s,a):=Q^\pi(s,a)-V^\pi(s)$. Since the distribution of returns can provide additional information about environmental dynamics, the distributional value function \cite{bellemare2017distributional, dabney2018implicit} is introduced to learn the distribution of accumulated rewards instead of its expectation's scalar value. For any state $s$, an analogous distributional Bellman equation of the form
\begin{equation}
    Z^{\pi}\left( s \right) :\xlongequal{D}r(s,a)+\gamma Z^\pi\left( S^\prime\right) 
\end{equation}
can be derived, where $Y\xlongequal{D}U$ indicates equality in distribution, that is the random variable $Y$ is distributed following the same law as $U$. The relationship between $V^\pi(s)$ and $Z^\pi(s)$ is given as follow
\begin{equation}
    V^\pi(s) = \mathbb{E}[Z^\pi(s)],
\end{equation}
which is the mean of the return distribution. According to the quantile representation in \cite{nam2021gmac,zhou2020non},  $Z(s)$ can be represented by a mixture of diracs from the sample values
\begin{equation}
    Z(s)\approx \tilde{Z}(s) :=\sum_{i=0}^{M-1}{\left( k _{i+1}-k _i \right) \delta _{d_i\left( s \right)}},
\end{equation}
where $0\le k_0\le k_1\le\cdots\le k_{M-1}\le1$ is a sequence of $M$ non-decreasing quantiles, each $d_i(s)$ is an estimation of the inverse CDF $F^{-1}_{Z(s)}(\hat{\tau}_i)$ corresponding to the quantile level $\hat{k}_i=(k_{i-1}+k_i)/2 $. 

\subsection{Low-level Controller of Actuation Motors}
We use the humanoid robot GR1, developed by Fourier Company \cite{fftai} (depicted in Figure \ref{fig: framework}). GR1 is a human-sized humanoid robot, standing 1.65 meters tall and weighing 55 kilograms. For whole-body control tasks, GR1 is operated through joint control torques $\tau\in\mathbb{R}^{23}$ at a frequency of 1000 Hz. It is important to note that while GR1 has a total of 44 degrees of freedom (DOFs), only 23 of these DOFs are utilized for locomotion tasks. The control torques are calculated based on the impedance control model
\begin{equation}
    \tau_t = K_p (q^{tgt}_t-q_t) + K_d(\dot{q}^{tgt}_t-\dot{q}_t) + \tau_{FF},
    \label{eq: pid}
\end{equation}
where $q^{tgt}_t$ represents the target position, $K_p$ and $K_d$ are the proportional-derivative (PD) parameters, and $\tau_{FF}$ refers to the feed-forward joint torques. 

Following prior work \cite{peng2017learning}, we set $\dot{q}^{tgt}_t$ and $\tau_{FF}$ as zeros to offer more stable training and better performance, leading to a simplification of (\ref{eq: pid}),
\begin{equation}
    \tau_t = K_P(q^{tgt}_t-q_t)-K_D\dot{q}_t.
    \label{eq: pd}
\end{equation}

\section{Learning Human-like locomotion}\label{metho}
\begin{figure}[thpb]
    \centering
    \includegraphics[scale=0.28]{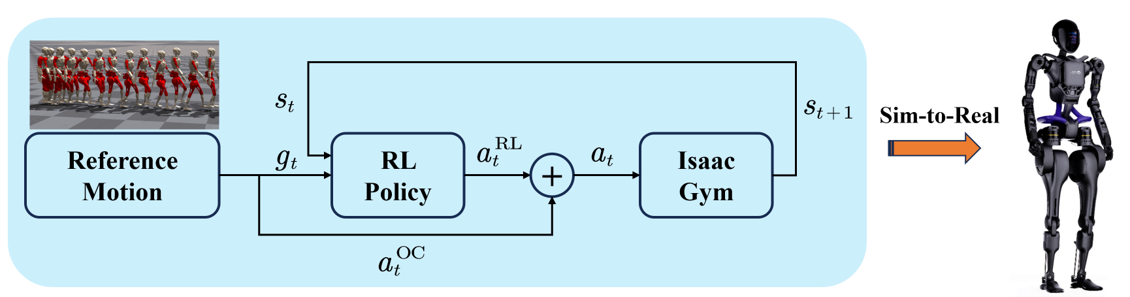}
    \caption{The overview of the proposed motion tracking pipeline. The RL policy takes the current state $s_t$ and the desired goal $g_t$ as inputs, and then outputs the RL action $a^{RL}_t$. This RL action is then combined with the open-loop control $a^{OC}_t$  to produce the final action $a_t$. The target joint position $a_t$ is subsequently converted into the torques applied to the joints.}
    \label{fig: framework}
  \end{figure}

In this section, we describe the training of an RL-based motion tracking controller within the Isaac Gym environment \cite{rudin2022learning} (see Figure \ref{fig: framework}). This controller produces human-like locomotion by minimizing the discrepancy between the current state of the robot and a walking reference motion. For successful sim-to-real transfer of the learned controller, a simple domain randomization method is proposed to perturb the dynamics of the training environment.

\subsection{Observation, Reward and Termination}
\textbf{Observation.} We represent the target walking motion as a sequence of target frames $\bar{s}_{1:T}$ with $T$ length. Each target frame $\bar{s}_t:=[\bar{v}_t,\bar{w}_t,\bar{q}_t,\dot{\bar{q}}]\in \mathbb{R}^{52}$ consists of the target linear and angular velocity of the pelvis, as well as the target positions and velocities of its actuating motors. At each step, the policy $\pi$ receives the current state of the humanoid robot defined as
\begin{equation}
    s_t=[\omega_t, g_p, q,\dot{q}_t,a_{t-1}]\in \mathbb{R}^{75},
\end{equation}
where $\omega_t$ represents the pelvis's angular velocity, $a_{t-1}$ is the last control command, and the projection of the gravity $g_p$ indicates the torso’s tilt relative to the gravity vector. Since the estimation of the pelvis's linear velocity in the real world is often accompanied by significant noise, we omit this information from $s_t$. To provide $\pi$ with the necessary information to track the target frame, the goal is given as 
\begin{equation}
    g_t=[\bar{q}_t,\bar{q}_t-q_t]\in\mathbb{R}^{46}.
\end{equation}
where the error term $\bar{q}_t-q_t$ is included to provide a redundant representation, similar to \cite{bergamin2019drecon}. By combining the current state with the goal, we create a 121-dimensional observation space. Previous studies \cite{peng2018deepmimic, bergamin2019drecon} utilized the 3D Cartesian positions and rotations of joints to assist physical characters in tracking a reference. However, we find that the joint positions are sufficient for the RL agent to make effective decisions.\\
\textbf{Reward Function.} To perform human-like locomotion while maintaining the stability of humanoid robots, the reward function $r(s_t,a_t,g_t)$ consists of the mimic reward and the behavioral regulation reward, i.e.,
\begin{equation}
    r(s_t,a_t,g_t) = r^m(s_t,a_t,g_t) + r^g(s_t,a_t,g_t).
\end{equation}
The mimic reward $r^{\text{m}}(s, a)$ measures the similarity between the robot's current state and corresponding target frame, which can be formulated as
\begin{equation}
    \begin{aligned}
        r^m_t=\exp(-\|q_t-\bar{q}_t\|^2_2),
    \end{aligned}
\end{equation}
\begin{figure}[thpb]
    \centering
    \includegraphics[scale=0.21]{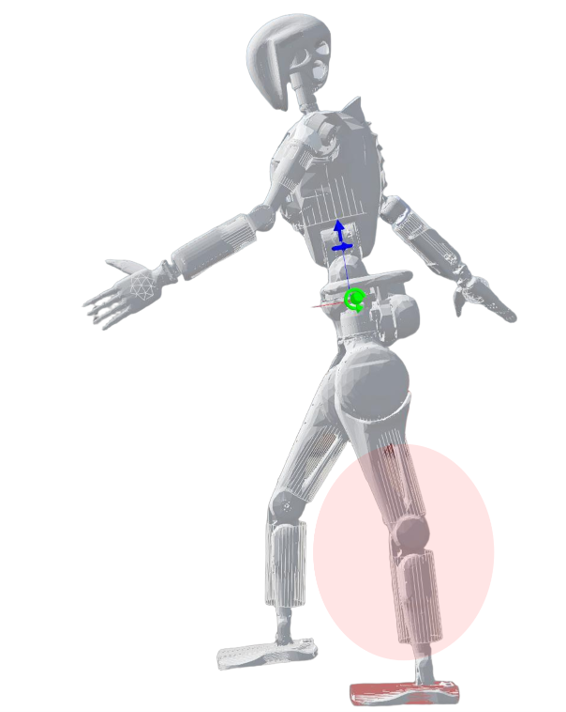}
    \caption{Without early termination, the humanoid robot tends to reach the maximum position of its actuation motors when tracking the reference.}
    \label{fig:undesirable}
  \end{figure}
Here we focus only on tracking joint positions rather than precisely following the velocity of the pelvis. This design stems from the challenge that remaps the pelvis's velocity from human movements to humanoid robots during motion retargeting. In traditional control methods like MPC, it is essential to optimize the target position of the pelvis based on the Zero Moment Point (ZMP) \cite{van2014zero} or other criteria \cite{zheng2010walking} to guarantee locomotion stability. We simplify this operation by eliminating the need for tracking the pelvis, which provides the RL agent with significant flexibility to maintain stability while tracking the reference.

The behavioral regulation reward $r^g_t$ is used to prevent undesirable behaviors, which can be decomposed into several sub-rewards
\begin{equation}
    r^g_t = r^{f}_t+ r^{smooth}_t+r^{energy}_t +0.5r^{osc}_t.
\end{equation}
To ensure long-term operation and energy efficiency, the foot reward $r^{f}_t$ imposes a penalty when the foot-ground contact exhibits excessive force or high velocities, 
\begin{equation}
    r^{f}_t=-\max\left(\frac{F^{foot}_t}{550} -1, 0\right) -\mathbb{I}(F^{foot}>0)\|v^{foot}_t\|^2_2,
    \label{eq:cone}
\end{equation}
where $F^{foot}_t$ represents the force applied to the ground, $\mathbb{I}(\cdot)$ is the indicator function, $v^{foot}_t$ is the velocity of the foot. The second term in (\ref{eq:cone}) serves a similar purpose to the friction cone, as discussed in \cite{caron2015stability} and \cite{fuchioka2023opt}, and it helps prevent slipping during locomotion. Additionally, the smooth reward $r^{smooth}_t = -0.5\|a_t-a_{t-1}\|^2_2$ imposes a penalty on non-smooth actions that could be difficult for the actuation motors to follow. The energy reward $r^{energy}_t=-0.1\|\tau\cdot \dot{q}\|$ can reduce the energy consumption during locomotion. To address the issue of the torso  oscillating and tilting while tracking reference motion, we introduce $r^{osc}_t$ as follows:
\begin{equation}
    r^{osc}_t=\exp(-30(1-|g_p^z|)),
\end{equation}
where $g^z_p$ is the gravity projected along the torso's z-axis. \\
\textbf{Early Termination}. In the context of cyclic human-like locomotion, typical early termination conditions involve detecting a fall.  A fall is identified when the robot's torso drops below a specific height. However, these criteria alone are insufficient for shaping the reward function to discourage undesirable behaviors, as shown in Figure \ref{fig:undesirable}. Since abnormal behaviors can potentially harm the robot by exceeding the maximum limitation of joint positions, we terminate an episode if any joint position exceeds 95$\%$ of its maximum limitation.

\subsection{Simple Domain Randomization}
\begin{figure}
    \centering
    \includegraphics[width=0.6\linewidth]{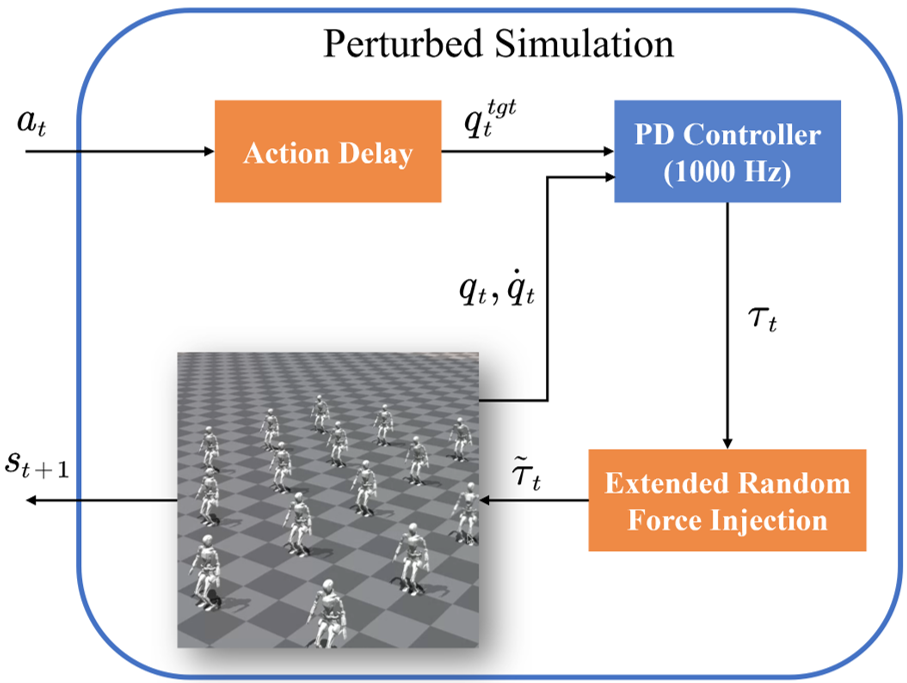}
    \caption{We propose a simple domain randomization method that incorporates ERFI and action delay into the training environment.}
    \label{fig:rnd}
\end{figure}

Classical domain randomization methods randomize various elements in the simulation, including the dynamic properties of legged robots, control parameters, and the physics of the environment. Additionally, these methods incorporate noise and time delays in the observations. As these methods rely on comprehensive expert knowledge, we propose simple domain randomization, achieved through the integration of ERFI and action delay, as shown in Figure \ref{fig:rnd}. This approach allows us to effectively perturb environmental dynamics with only a few lines of code. A comprehensive comparison with more complex domain randomization techniques is presented in Section \ref{reseult}.

Based on previous work \cite{campanaro2024learning} that has demonstrated the effectiveness of ERFI in the domain of quadruped robots, we apply similar perturbations to the joints of humanoid robots. Specifically, we randomized the forces applied to the actuation motors of GR1 at a frequency of 1000 Hz, as expressed below:
\begin{equation}
    \tilde{\tau}_t = \tau_t + (1-m^o)\tau^{RFI}_t+ m^o\tau^{RAO},
\end{equation}
where the random force injection (RFI) $\tau^{RFI}_t\in\mathbb{R}^{23}$ is sampled at each simulation step. Meanwhile, the random actuation offset (RAO) $\tau^{RAO}_t\in\mathbb{R}^{23}$ and the mask $m^o\in \{0,1\}$ are sampled at the beginning of each episode, respectively. Here we sample $m^o$ following a Bernoulli distribution, with $m^o\sim Bernueli(0.5)$. This indicates that we utilize RFI with 50$\%$ of the parallelized RL training environments, while the remaining 50$\%$ rely on RAO. 

The actuation dynamics is implicitly randomized by $\tau^{RFI}_t$, which is sampled from a uniform distribution given by
\begin{equation}
    \tau^{RFI}_t \sim \mathcal{U}(-\alpha \tau^{lim}, \alpha \tau^{lim}),
\end{equation}
where $\alpha\in(0, 1)$ is a scalar, $\tau^{lim}\in\mathbb
{R}^{23}$ refers to the maximum torque of the actuation motors. In contrast, $\tau^{RAO}$ implicitly models offsets in the joint position or in the payload supported by the robot and is sampled at the beginning of each episode
\begin{equation}
    \tau^{RAO} \sim \mathcal{U}(-\alpha\tau^{lim},\alpha\tau^{lim}).
\end{equation}
Here $\alpha$ reflects the discrepancy between the simulation and the real world. A large $\alpha$ indicates a higher level of perturbation in environmental dynamics, making the RL policy more robust. However, this also increases the non-smoothness of the actions, making it more challenging for the actuation motors to respond effectively. To balance the robustness and smoothness of the RL policy, we set $\alpha$ to $0.1$ in our experiments, which means that we assume a maximum difference of 10$\%$ between the RL training environment and the real world.

\begin{figure}[thpb]
    \centering
    \includegraphics[scale=0.25]{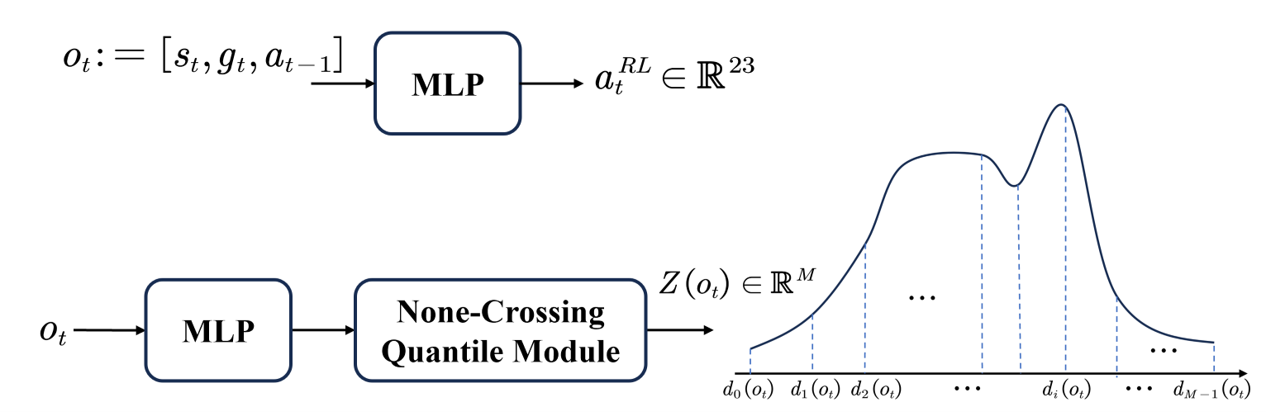}
    \caption{Neural networks are used to parameterize the policy and the distributional value function. MLP refers to the multi-layer perceptron, the non-crossing quantile module has the same architecture as that of previous work \cite{zhou2020non}.}
    \label{fig:nn}
  \end{figure}

While ERFI is effective at randomizing environmental dynamics, it fails to account for the response delay of the actuation motors when the robot's lower body is under high load conditions. To address this action delay, the control signal sent to the PD controller is calculated as follows
\begin{equation}
    q^{tgt}_t=(1-\beta)q^{tgt}_{t-1}+\beta a_t,\quad \beta\sim \mathcal{U}(0,1)
\end{equation}
where $q^{tgt}_0$ is initialized as a zero vector.

\subsection{Action, Network and Optimization Objective}
As the reference motion details the gaits of humanoid robots, we decouple the locomotion controller into two parts. One is an open-loop control part that determines the target joint positions at each time step. The other is an RL-based feedback part that adjusts the joint positions on top of the reference trajectory based on the observation $[s_t,g_t]$,
\begin{equation}
    a_t = a^{OC}_t+a^{RL}_t, \quad a^{RL}_t \sim \pi(a^{RL}_t|s_t,g_t),
    \label{eq:comb}
\end{equation}
where $a^{OC}_t:=\bar{q}_t$ is the open-loop component and $\pi(a^{RL}_t|s_t,g_t)$ is the feedback component. The policy's action distribution $\pi(a^{RL}_t|s_t,g_t)$ is represented using a multi-dimensional Gaussian with a fixed diagonal covariance matrix $\sigma^\pi=\exp(-2.9)$. 

This decomposition in (\ref{eq:comb}) provides a convenient way to adjust the motion pattern of humanoid locomotion. Specifically, we can further decompose the control action $a_t$ into the left lower body, the right lower body, and the upper body
\begin{equation}
    \begin{aligned}
        a_t=&\left[w_1\bar{q}_t^{lower,l}, w_2\bar{q}_t^{lower,r},w_3\bar{q}^{upper}_t\right]^T + \\
            & \left[a^{RL,lower,l}_t, a^{RL,lower,r}_t,a^{RL,upper}_t\right]^T
    \end{aligned}
    \label{eq:combine}
\end{equation}
where the weight $w:=[w_1,w_2$, $w_3]$ can be modified based on the specific task at hand. During the training process, we set $w_1=w_2=w_3=1$. 

A schematic illustration of the policy and the distributional value networks is available in Figure \ref{fig:nn}. The inputs to each network consist of the current state $s_t$, the goal $g_t$, and previous action $a_{t-1}$. We update the parameters $\psi$ of the policy network by minimizing the following objective:
\begin{equation}
    \psi \leftarrow \arg\min \mathcal{L}^{actor}_{RL}(\psi) + 10\mathcal{L}_{symmetric}(\psi) 
    \label{eq:loss}
\end{equation}
where $\mathcal{L}^{actor}_{RL}(\psi)$ is the policy loss function proposed in \cite{schulman2017proximal}, and the mirror loss is integrated to ensure that the controller learns more symmetric locomotion behavior. Specifically, we utilize the mirror loss proposed by \cite{yu2018learning}:
\begin{equation}
    \mathcal{L}_{symmetric}(\psi)=\|a^{RL}_t-\Phi_a(a^{RL,sym}_t)\|^2,
\end{equation}
where $\Phi_a$ and $\Phi_s$ mirror the state and action along the humanoid robot's sagittal plane, $a^{RL}_t\sim \pi(a^{RL}_t|s_t,g_t)$, $a^{RL,sym}_t\sim \pi(a^{RL,sym}_t|\Phi_s(s_t, g_t))$. The distributional value function $Z(s,g|\phi)$ parameterized with parameters $\phi$ is updated according to the Wasserstein metric \cite{dabney2018distributional},
\begin{equation}
    \phi \leftarrow \arg\min \frac{1}{2M}\sum^{M-1}_{i=0}\sum^{M-1}_{j=0}\left|\tau - \mathbb{I}(\sigma_{ij})<0\right|\sigma^2_{ij},
    \label{eq:value}
\end{equation}
where $\sigma_{ij}=r(s,a,g) + \gamma d_j(s^\prime,g^\prime)-d_i(s,g)$ is the distributional TD error.

\section{Experimental Evaluation }\label{reseult}
In this section, we conduct comprehensive experiments to answer the following questions: (1) Can the motion tracking controller learn human-like locomotion and be robust to external perturbations? (2) Can the motion patterns of humanoid robots be modified through the residual mechanism without fine-tuning? (3) Can the proposed technique of integrating ERFI with action delay serve as an alternative to complex domain randomization? (4) Can the distributional value estimation be more beneficial than commonly used value estimation methods in RL-based humanoid locomotion? To answer these questions, we evaluate the proposed framework in high-fidelity simulations and real-world systems. All results are illustrated in the associated \href{https://www.bilibili.com/video/BV1nBPye8Exd/?spm_id_from=333.337.search-card.all.click}{video}.
\subsection{Training Details}
To create the reference walking motion, we apply the motion retargeting operator to map human walking data to that of the robot GR1. In the reference motion, GR1 moves forward at a speed of 0.5 m/s for 2 seconds after standing still for 1 second, and finally maintains a standing pose. The motion tracking controller operates at a frequency of 50 Hz, while the PD controller and the Isaac simulation run at 1000 Hz. The motion tracking controller is trained for approximately 3 hours in Isaac Gym using 4096 parallel environments with a single RTX 4090. For updating the policy and distributional value networks, we collect a maximum of 204,800 samples at each update. The update is done by performing stochastic sub-gradient descent on the loss functions (\ref{eq:loss}) and (\ref{eq:value}). We use a batch size of 5120 and perform 8 updates using the Adam optimizer. The step size for Adam is set to $1e-4$ for both networks during the training process. 

\begin{figure*}
\centering 
\subfigure[Normal walking]
{\includegraphics[width=0.35\textwidth]{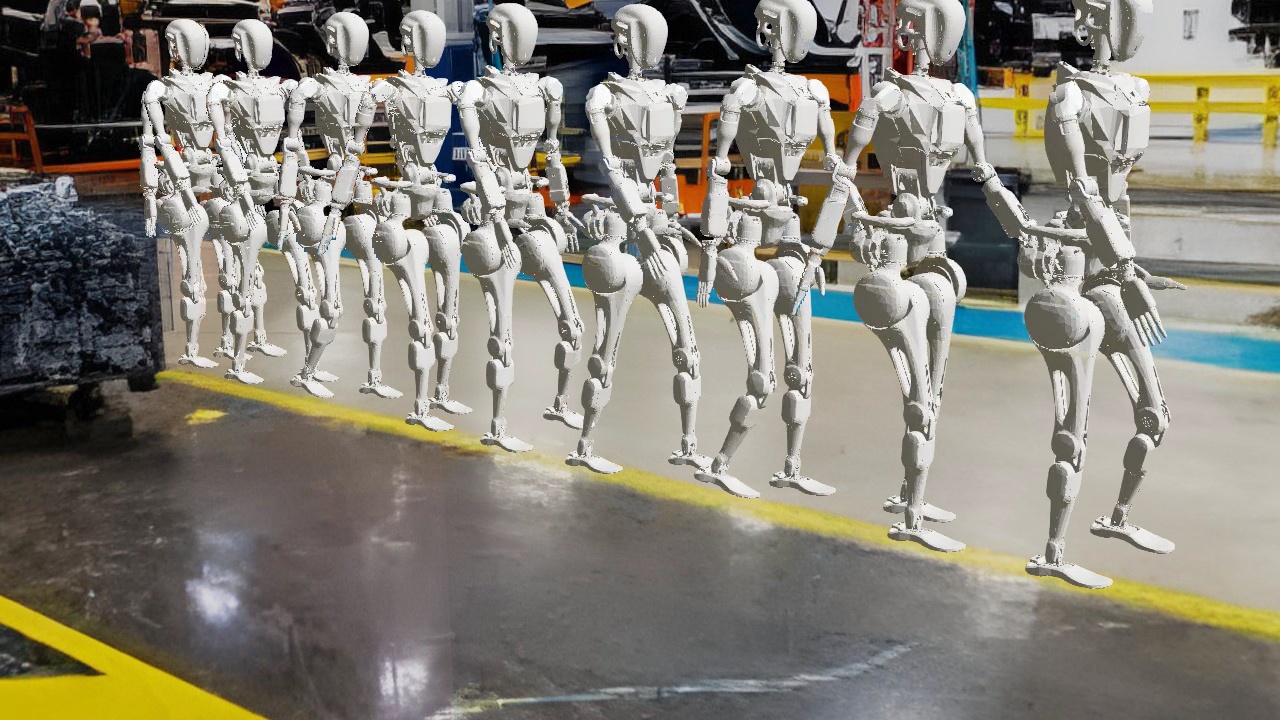}}
\subfigure[The control signal of the hip joint]
{\includegraphics[width=0.3\textwidth]{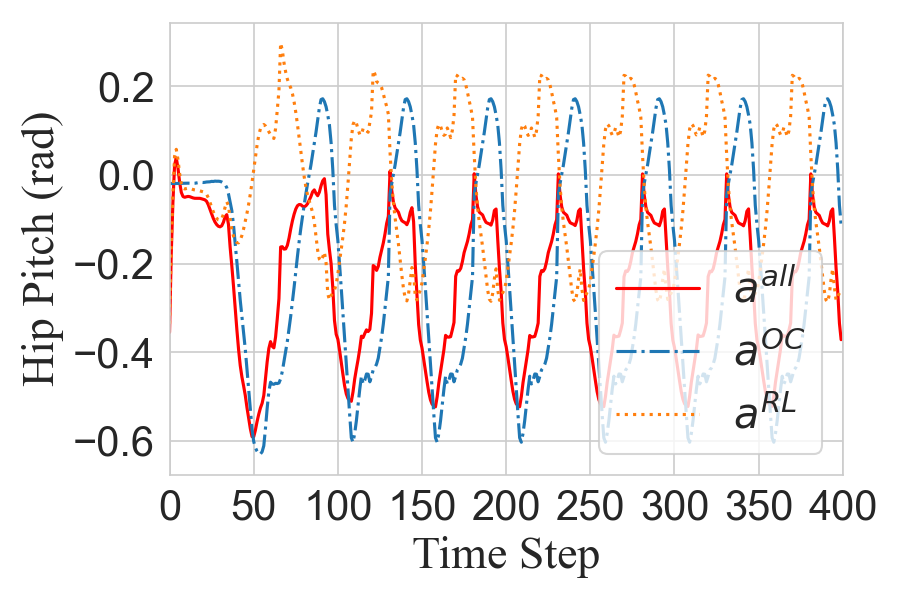}}
\subfigure[The control signal of the knee joint]
{\includegraphics[width=0.3\textwidth]{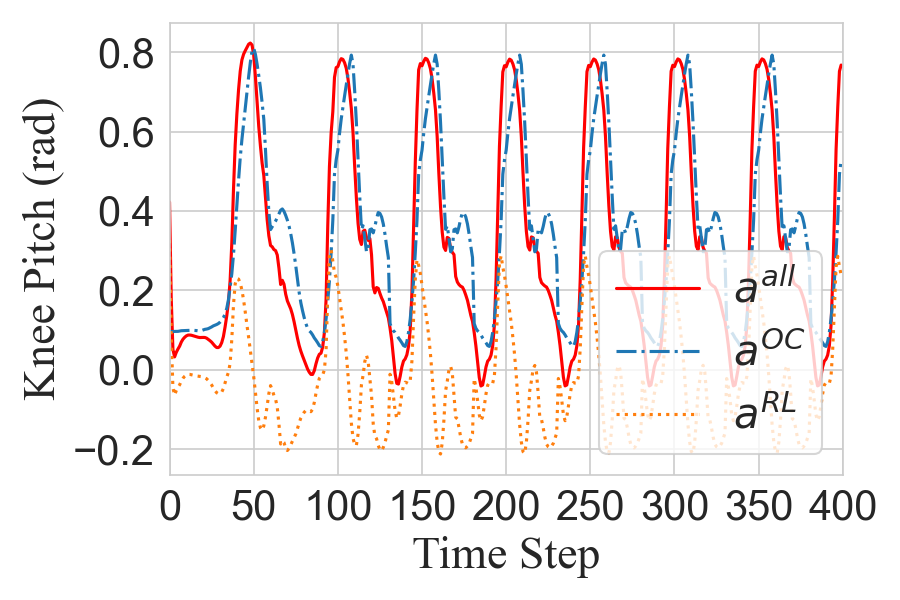}}
\caption{Performance of the motion tracking controller on the high-fidelity simulation Webots.}
\label{fig:walking}
\end{figure*}

\begin{figure*}
\centering 
\subfigure[Walking forward while shifting to the left]
{\includegraphics[width=0.32\textwidth]{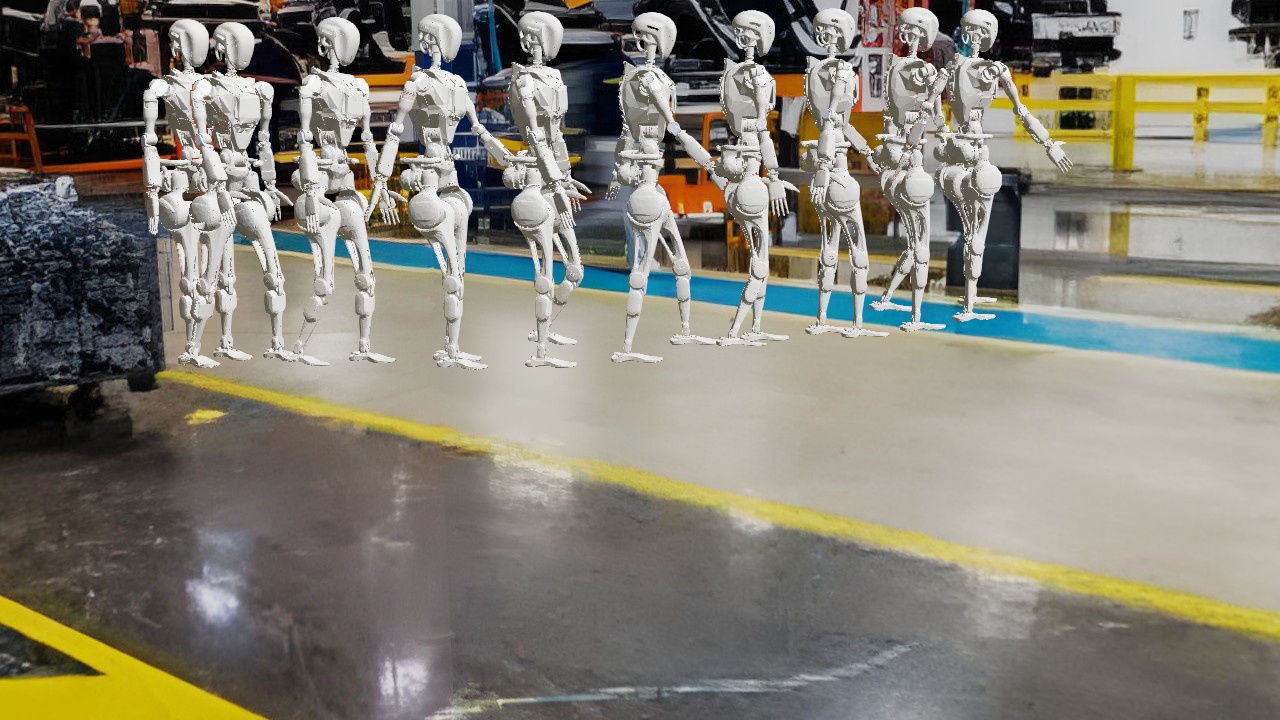}}
\subfigure[Walking forward while shifting to the right]
{\includegraphics[width=0.32\textwidth]{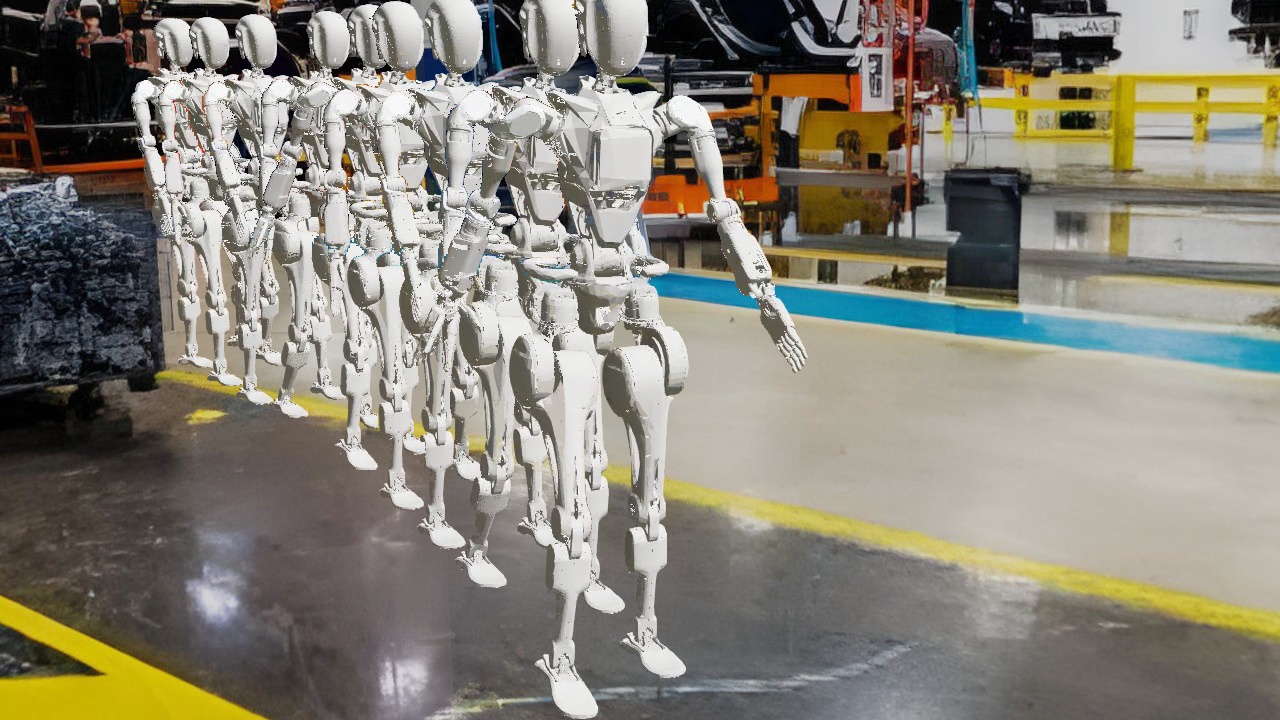}}
\subfigure[Slight swings of the arms]
{\includegraphics[width=0.32\textwidth]{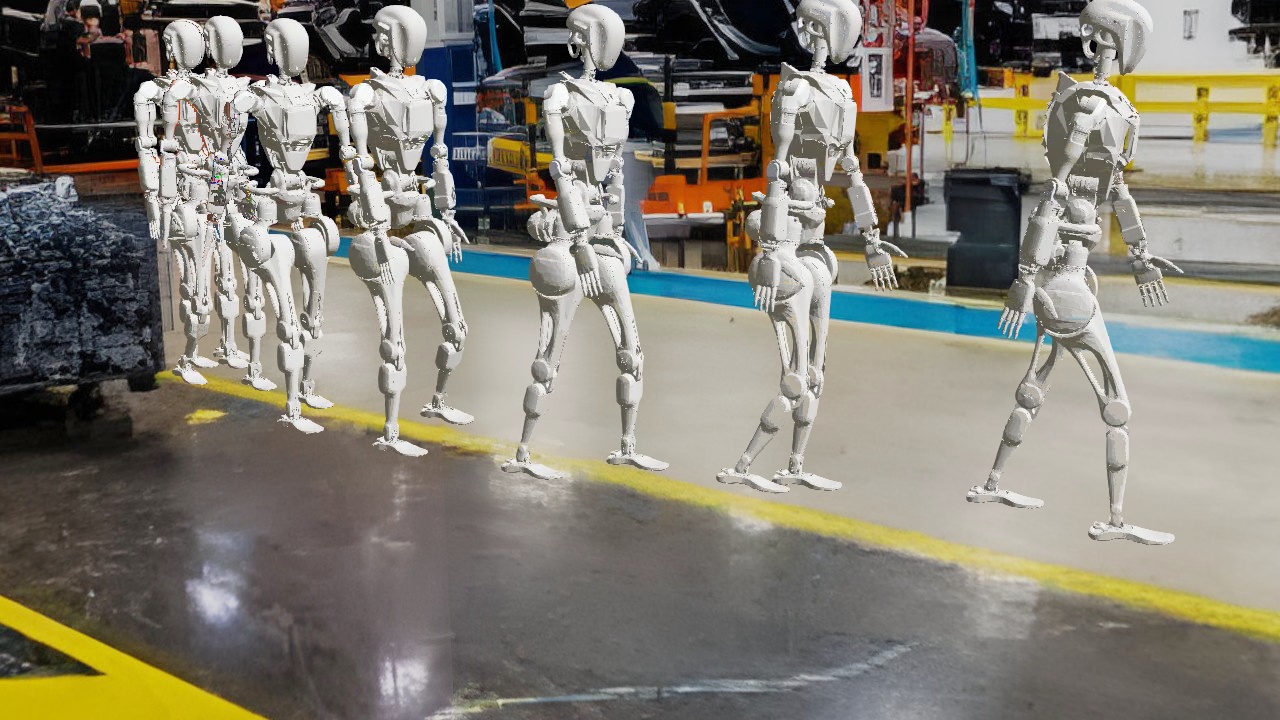}}
\caption{Adaptation of humanoid motion patterns without fine-tuning. By using different combinations of the open-loop control and the RL policy, the humanoid robot exhibits three distinct motion patterns.}
\label{fig:adapt}
\end{figure*}

\begin{figure*}
\centering 
\subfigure[Normal walking]
{\includegraphics[width=0.32\textwidth]{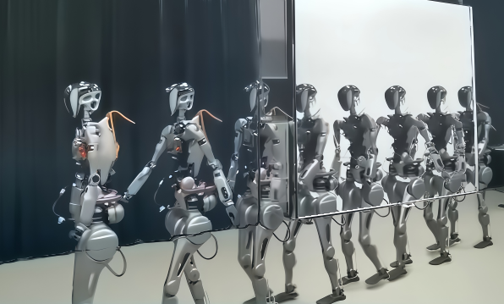}\label{real:a}}
\subfigure[Walking forward while shifting to the left]
{\includegraphics[width=0.32\textwidth]{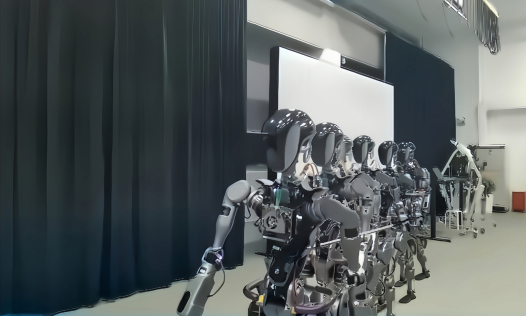}\label{real:b}}
\subfigure[External force perturbation]
{\includegraphics[width=0.32\textwidth]{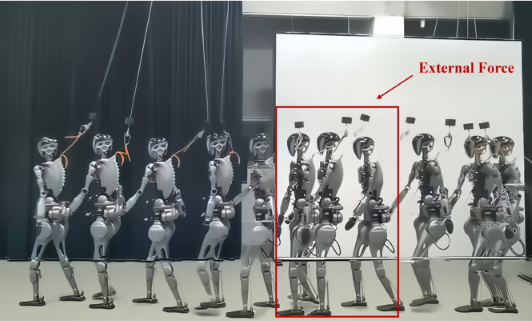}\label{real:c}}
\caption{The learned motion tracking controller can successfully transfer from simulation to the real world.}
\label{fig:real}
\end{figure*}

\subsection{Performance on High Fidelity Simulation}
\textbf{Sim-to-Sim Verification.} We evaluate the learned controller using Webots, which offers a highly realistic physical simulation environment. Given the cyclic nature of locomotion, we can continuously send walking segments sampled from the reference motion to the learned controller. This allows the humanoid robot to move continuously, as shown in Figure \ref{fig:walking}. The humanoid robot exhibits human-like gait patterns, including natural foot placement and body posture adjustments. Moreover, due to early termination, the gait of the humanoid robot is also shaped to avoid reaching the extreme position of its actuation motors.

\textbf{Motion Pattern Adjustment.} Human-like locomotion is crucial for human-centric scenarios, but different motion patterns are needed to meet various requirements of tasks in real applications. Instead of training the motion tracking controller to track different reference motions from scratch, we can simply adjust the weight $w$ in (\ref{eq:combine}) to modify the motion patterns of humanoid robots. As illustrated in Figure \ref{fig:adapt}, the weight is set to $[0.8, 1.2, 1.0]$ for the shift to the left, $[1.1, 0.8, 1.0]$ for the shift to the right, and $[0.9, 1.2, 0.6]$ for slight swings of the arms, respectively. Since the swing amplitudes of the left and right legs in the reference motion are not identical, the weights used for left and right shifts produce asymmetric results.
In addition, we set the swing amplitude of the arms at 60$\%$ of that in the training environment, which required adjustments to the weights $w_1$ and $w_2$ for the lower body to accommodate the slight swings of the arms. From a temporal perspective, the adaptation of motion with zero-shot transfer can be attributed to the joint control command $a_t$, which resembles sinusoidal wave signals during locomotion tasks. Even when we alter the magnitude of the open-loop action $a^{OC}_t$, this action maintains synchronization in phase with the combined action $a_t$, ensuring coordinated and stable movements. It is important to note that this motion adjustment is also beneficial when the motion tracking controller is deployed in the real world. Due to assembly errors or variations in the motor performance of the robot's left and right legs, asymmetric movements might hinder the humanoid robot's ability to walk straight. However, by adjusting the weights in (\ref{eq:combine}), we can achieve the desired motion effectively.

\subsection{Performance on Real System}
We conduct evaluations in the real world to verify the transferability of the learned controller. We set the weight $w$ in (\ref{eq:combine}) to $[1, 1, 1]$ for the normal walking task. As depicted in Figure \ref{real:a},  the gait of the humanoid robot closely matches that of the simulation. To perform the left shift task illustrated in \ref{real:b}, we adjust the weight $w$  to $[0.9, 1., 1]$. This change in $w$ reflects slight differences in dynamics between the simulation environment and the real world. Moreover, we tested the robustness of the learned RL policy against external force perturbations. Figure \ref{real:c} demonstrates that our policy effectively resists such disturbances, proving its efficacy and reliability.

\begin{figure*}
\centering 
\subfigure[ ]
{\includegraphics[width=0.32\textwidth]{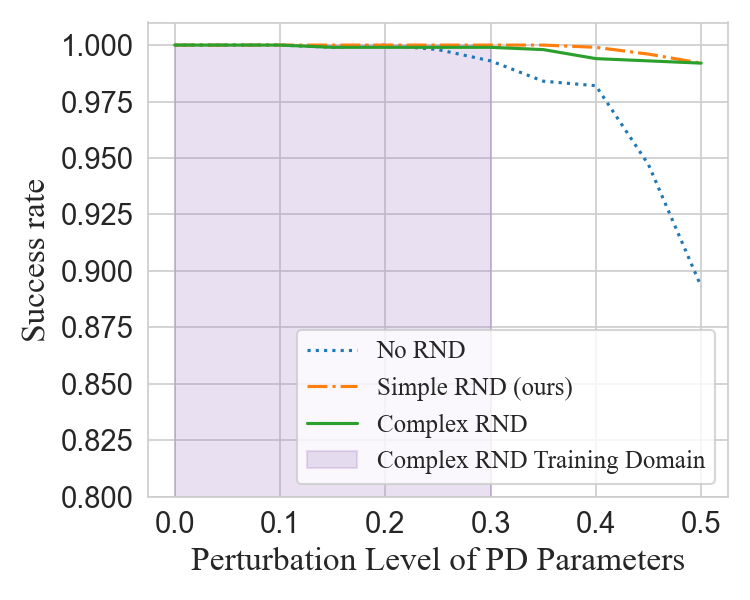}}
\subfigure[ ]
{\includegraphics[width=0.32\textwidth]{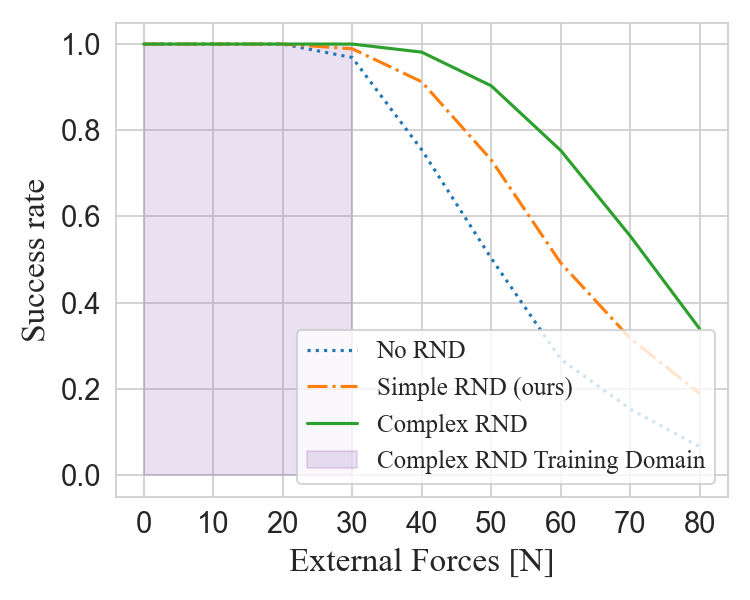}\label{rnd:a}}
\subfigure[ ]
{\includegraphics[width=0.32\textwidth]{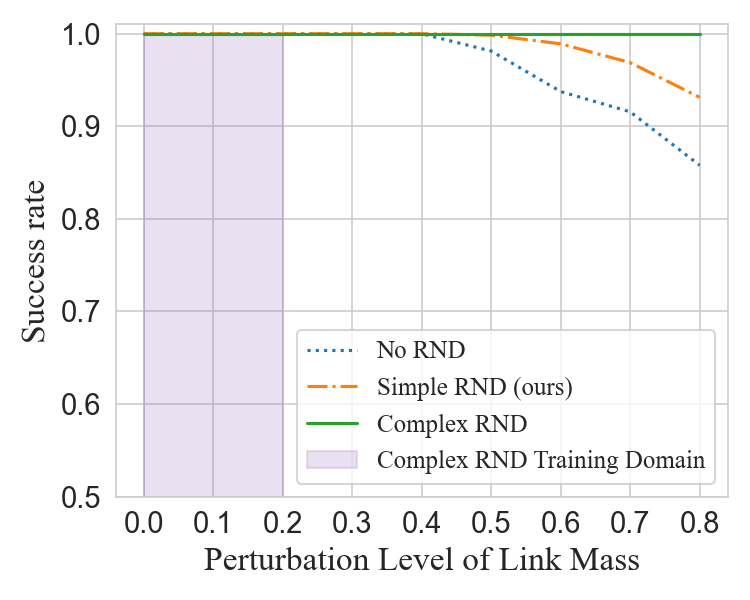}}
\caption{Comparison of Simple RND, Complex RND, and No RND based on variations in PD parameters, external forces, and link mass.}
\label{fig:rnd_compare}
\end{figure*}

\subsection{Ablation Study}
We perform ablation on the two main ingredients proposed in this work: 1) simple domain randomization, and 2) distributional-based value estimation. 

\textbf{Simple Domain Randomization.} We compare our proposed simple domain randomization (RND) with complex RND and No RND. For more details on complex RND, please refer to \cite{radosavovic2024real}. The metric used to evaluate their performances is the success rate at which the humanoid robot can track the reference motion without violating the constraints applied in early termination. Figure \ref{fig:rnd_compare} illustrates the robustness of different RND methods in response to changes in the link mass, the application of external forces, or variations in the PD parameters of the actuation motors. The complex RND method perturbs the PD parameters of the controller, the magnitude of the control torques and the damping force to improve the policy's robustness to motors. However, as shown in Figure \ref{rnd:a}, the simple RND can also enhance the policy's robustness to motor variations using only ERFI and action delay. We set the only hyperparameter $\alpha$ of the simple RND to 0.1, indicating that the maximum external force perturbation applied to the joints during training is 10$\%$ of the maximum torque. In comparison with the complex RND, which is resilient to various levels of perturbations (as shown in Figure \ref{fig:rnd_compare}), the simple RND is primarily robust only to moderate perturbations. While we could increase the robustness of simple RND to larger perturbations by raising $\alpha$, in practice, a value of $\alpha = 0.1$ is sufficient for sim-to-real transfer. Increasing $\alpha$ further would reduce the smoothness of the policy output, which could undermine the effectiveness of sim-to-real transfer.

\begin{figure}
    \centering
    \includegraphics[width=0.9\linewidth]{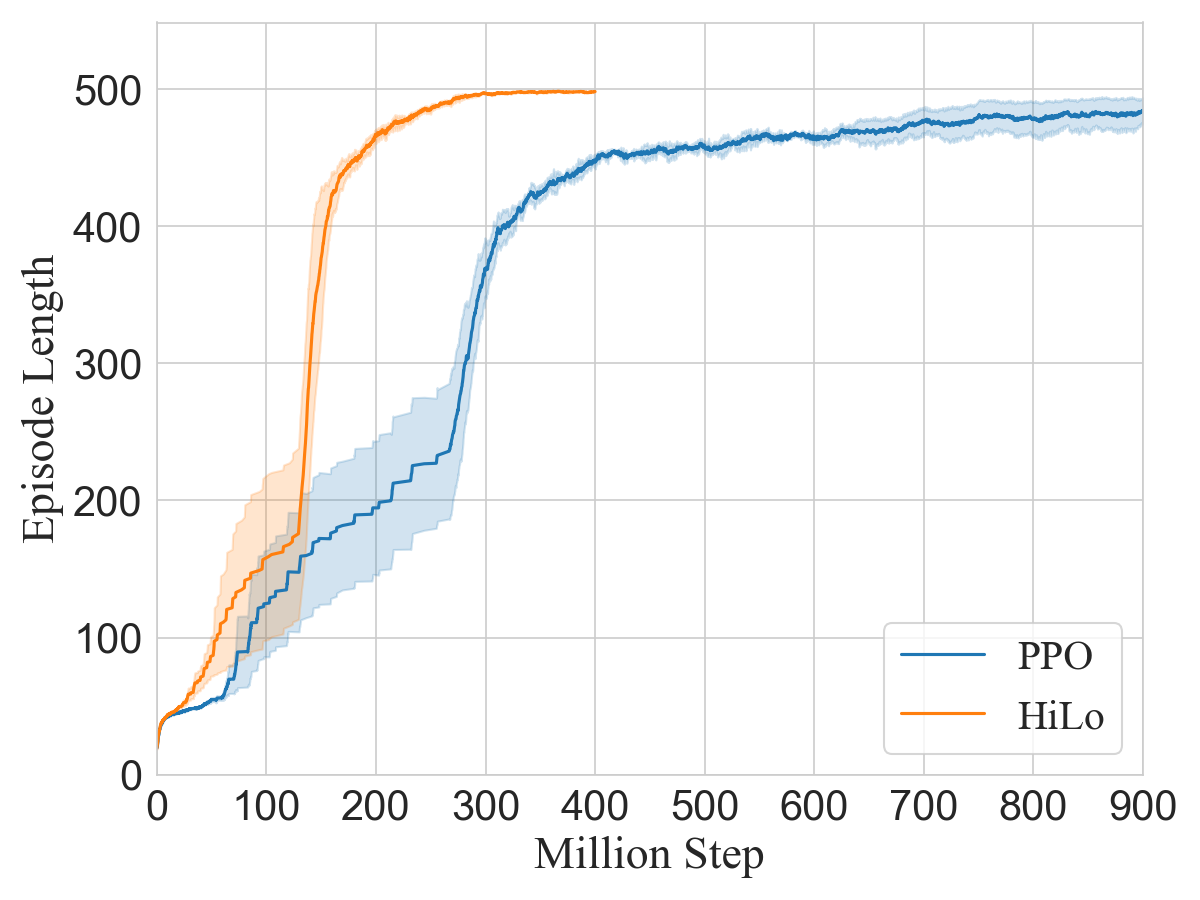}
    \caption{Learning curves. All curves are the average of three runs with different random seeds, and the shaded areas are the standard deviation.}
    \label{fig:lr}
\end{figure}

\textbf{Distributional Value Estimation.} The only difference between our proposed RL algorithm and the commonly used Proximal Policy Optimization (PPO) algorithm \cite{schulman2017proximal} is that we employ the distributional value function. We compare our method with PPO to evaluate the effectiveness of our distributional value estimation. Both algorithms are trained using 3 different seeds within the perturbed environment whose dynamics are randomized by our proposed simple RND. Due to early termination, the length of an episode can directly reflect how long the humanoid robot can maintain balance, avoid undesirable behaviors, and track the reference motion.  Therefore, we use episode length as the performance metric to compare the different algorithms. The learning curves shown in Figure \ref{fig:lr} indicate that our distributional value estimation achieves convergence 2.5 times faster than PPO. 

\section{Conclusion}\label{concul}
In this work, we present HiLo, an effective framework for humanoid robots to achieve human-like locomotion by tracking reference motions. By integrating a residual mechanism, simple domain randomization, and a distributional value function, HiLo effectively trains an RL-based motion tracking controller. The learned controller allows the humanoid robot to perform natural and agile gaits while exhibiting resilience to external disturbances in real-world environments. Furthermore, the motion patterns of the humanoid robots can be modified using the residual mechanism without the need for fine-tuning. Our comprehensive experiments highlight the feasibility and effectiveness of leveraging the motion tracking controller to achieve human-like locomotion and task adaptation, making it a promising solution to enhance the capabilities of humanoid robots in human-centric scenarios. In future work, we plan to incorporate multiple reference trajectories to improve the adaptability and versatility of the learned controller. Additionally, we will introduce a motion generator that can produce target references based on joystick input.

\bibliographystyle{IEEEtran}
\bibliography{IEEEabrv,reference}

\end{document}